\documentclass[pdflatex,sn-apa]{sn-jnl}
\usepackage{times}
\begin{document}






\section{MLJ Contribution Information Sheet}

\textbf{Title:} Classifier Calibration -- How to assess and improve predicted class probabilities: a survey

\ \\
\emph{\textbf{Q1: What is the main claim of the paper? Why is this an important contribution to the machine learning literature?}}

We present an extensive survey on the topic of calibration of multiclass probability estimates which is missing in the literature since the works of \citet{lichtenstein1976,niculescu-mizil2005}, which focused on the binary case and are missing recent advancements.  
Being able to quantify the correctness of posterior probability estimates is crucial in critical applications, decision making, cost-sensitive classification and autonomous systems (e.g. medical diagnosis, self-driving cars).
The importance of this field is clearly visible in the recent increase in publications in the areas of classifier calibration, uncertainty estimation, conformal prediction, with a recent focus on reducing the overconfidence of state-of-the-art deep neural networks.
This interest has led to several recent advances in methods to improve and evaluate calibration.
However, there is no current reference for new practitioners to easily access all the main available calibration tools, which occasionally leads to misunderstandings in the evaluation of calibration, or the use of misleading visualisations.
This paper gives a general overview of the field from the early stages, providing context and the evolution of evaluation measures, multiple visualisation techniques to inspect the quality of probability estimations, multiple post-hoc calibration methods with an comprehensive comparison with carefully designed synthetic datasets, and related advanced topics which point to opportunities for future research.

\ \\
\noindent
\emph{\textbf{Q2: What is the evidence you provide to support your claim? Be precise.}}

We provide a detailed motivation about the importance of the field of classifier calibration. We introduce the reader to the topic with a tutorial which covers the most important concepts in a simplified manner.
We then give precise formal definitions of concepts, providing multiple toy examples that help the reader understand each idea. We also compare methods on carefully selected synthetic data which demonstrate the differences between them.
In addition, all of the sections are accompanied by many graphs which clarify every part of the survey. Finally, most of our references were published in 2005 or later, that is after the work of \citet{niculescu-mizil2005}, supporting our claim that our work discusses recent advancements in the field that were not covered in previous surveys on the subject of calibration.

\ \\
\noindent
\emph{\textbf{Q3: What papers by other authors make the most closely related contributions, and how is your paper related to them?}}

This is the most recent survey about classifier calibration since the now outdated work of \citet{lichtenstein1976} which only covered the binary case, and which could not include recent work that has appeared in the last decades.
The most recent analysis of binary calibration in machine learning was presented by \citet{niculescu-mizil2005}.
Other related papers include the following. 

\citet{mena2021} give a recent survey on the estimation of uncertainty from a Bayesian perspective, focusing on Deep Neural Networks.
The emphasis is on the description of the different types of uncertainty presented by various state-of-the-art models, being one of the components of probabilistic calibration.
However, the section on calibration is brief and only mentions the existence of the basic form of confidence calibration, along with line reliability diagrams which we extend in this survey.

\citet{hullermeier2021aleatoric} give an introduction to the concepts of aleatoric and epistemic uncertainty from a theoretical perspective.
The authors also analyse the sources of uncertainty in supervised learning, and deepen the analysis for a multitude of common classification models.
Calibration (which broadly concerns aleatoric uncertainty) is mentioned briefly as a related research topic without further analysis or discussion.

The most recent work on evaluation of classifier calibration can be found in \citet{vaicenavicius2019evaluating,widmann2019,kull2019}. 
All of them propose new evaluation measures and/or calibration methods with different notations.
In our survey we review their findings and highlight their individual contributions.

\citet{brocker2007} summarise the evolution of common calibration visualisation tools, and propose novel reliability diagrams.
In our survey, we built on this summary and expanded it with more recent work.

\ \\
\noindent
\emph{\textbf{Q4: Have you published parts of your paper before, for instance in a conference?}}

Part of this survey has been delivered in the form of a tutorial that we presented at the 2020 European Conference on Machine Learning and Principles and Practice of Knowledge Discovery in Databases (see \url{https://classifier-calibration.github.io}).
We extend the work by stating the importance of classifier calibration and highlighting existing work that hasn't been credited and which proved to be crucial for later advances in classifier calibration.

\bibliography{bibliography}

\begin{thebibliography}{}
\providecommand{\doi}[1]{\url{https://doi.org/#1}}
\bibcommenthead

\bibitem [\protect \citeauthoryear {%
Allikivi%
\ \BBA {} Kull%
}{%
Allikivi%
\ \BBA {} Kull%
}{%
{\protect \APACyear {2019}}%
}]{%
allikivi2019}
\APACinsertmetastar {%
allikivi2019}%
\begin{APACrefauthors}%
Allikivi, M\BHBI L.%
\BCBT {}\ \BBA {} Kull, M.%
\end{APACrefauthors}%
\unskip\
\newblock
\APACrefYearMonthDay{2019}{}{}.
\newblock
{\BBOQ}\APACrefatitle {{Non-parametric Bayesian Isotonic Calibration: Fighting
  Over-confidence in Binary Classification}} {{Non-parametric Bayesian Isotonic
  Calibration: Fighting Over-confidence in Binary Classification}}.{\BBCQ}
\newblock
 \APACrefbtitle {Joint European Conference on Machine Learning and Knowledge
  Discovery in Databases ({ECML-PKDD'19})} {Joint european conference on
  machine learning and knowledge discovery in databases ({ECML-PKDD'19})}\
  (\BPGS\ 68--85).
\PrintBackRefs{\CurrentBib}

\bibitem [\protect \citeauthoryear {%
Angelopoulos%
\ \BBA {} Bates%
}{%
Angelopoulos%
\ \BBA {} Bates%
}{%
{\protect \APACyear {2021}}%
}]{%
angelopoulos2021gentle}
\APACinsertmetastar {%
angelopoulos2021gentle}%
\begin{APACrefauthors}%
Angelopoulos, A.N.%
\BCBT {}\ \BBA {} Bates, S.%
\end{APACrefauthors}%
\unskip\
\newblock
\APACrefYearMonthDay{2021}{}{}.
\newblock
{\BBOQ}\APACrefatitle {A gentle introduction to conformal prediction and
  distribution-free uncertainty quantification} {A gentle introduction to
  conformal prediction and distribution-free uncertainty
  quantification}.{\BBCQ}
\newblock
\APACjournalVolNumPages{arXiv preprint arXiv:2107.07511}{}{}{}.
\newblock

\newblock

\PrintBackRefs{\CurrentBib}

\bibitem [\protect \citeauthoryear {%
Ayer%
, Brunk%
, Ewing%
, Reid%
\BCBL {}\ \BBA {} Silverman%
}{%
Ayer%
\ \protect \BOthers {.}}{%
{\protect \APACyear {1955}}%
}]{%
ayer1955empirical}
\APACinsertmetastar {%
ayer1955empirical}%
\begin{APACrefauthors}%
Ayer, M.%
, Brunk, H.D.%
, Ewing, G.M.%
, Reid, W.T.%
\BCBL {} Silverman, E.%
\end{APACrefauthors}%
\unskip\
\newblock
\APACrefYearMonthDay{1955}{}{}.
\newblock
{\BBOQ}\APACrefatitle {An empirical distribution function for sampling with
  incomplete information} {An empirical distribution function for sampling with
  incomplete information}.{\BBCQ}
\newblock
\APACjournalVolNumPages{The annals of mathematical statistics}{}{}{641--647}.
\newblock

\newblock

\PrintBackRefs{\CurrentBib}

\bibitem [\protect \citeauthoryear {%
Barlow%
\ \BBA {} Brunk%
}{%
Barlow%
\ \BBA {} Brunk%
}{%
{\protect \APACyear {1972}}%
}]{%
barlow1972isotonic}
\APACinsertmetastar {%
barlow1972isotonic}%
\begin{APACrefauthors}%
Barlow, R.E.%
\BCBT {}\ \BBA {} Brunk, H.D.%
\end{APACrefauthors}%
\unskip\
\newblock
\APACrefYearMonthDay{1972}{}{}.
\newblock
{\BBOQ}\APACrefatitle {The isotonic regression problem and its dual} {The
  isotonic regression problem and its dual}.{\BBCQ}
\newblock
\APACjournalVolNumPages{Journal of the American Statistical
  Association}{67}{337}{140--147}.
\newblock

\newblock

\PrintBackRefs{\CurrentBib}

\bibitem [\protect \citeauthoryear {%
Bengs%
, H{\"u}llermeier%
\BCBL {}\ \BBA {} Waegeman%
}{%
Bengs%
\ \protect \BOthers {.}}{%
{\protect \APACyear {2022}}%
}]{%
bengs2022pitfalls}
\APACinsertmetastar {%
bengs2022pitfalls}%
\begin{APACrefauthors}%
Bengs, V.%
, H{\"u}llermeier, E.%
\BCBL {} Waegeman, W.%
\end{APACrefauthors}%
\unskip\
\newblock
\APACrefYearMonthDay{2022}{}{}.
\newblock
{\BBOQ}\APACrefatitle {Pitfalls of Epistemic Uncertainty Quantification through
  Loss Minimisation} {Pitfalls of epistemic uncertainty quantification through
  loss minimisation}.{\BBCQ}
\newblock
 A.H.~Oh, A.~Agarwal, D.~Belgrave\BCBL {}\ \BBA {} K.~Cho\ (\BEDS),
  \APACrefbtitle {Advances in Neural Information Processing Systems.} {Advances
  in neural information processing systems.}
\PrintBackRefs{\CurrentBib}

\bibitem [\protect \citeauthoryear {%
Brier%
}{%
Brier%
}{%
{\protect \APACyear {1950}}%
}]{%
brier1950}
\APACinsertmetastar {%
brier1950}%
\begin{APACrefauthors}%
Brier, G.W.%
\end{APACrefauthors}%
\unskip\
\newblock
\APACrefYearMonthDay{1950}{}{}.
\newblock
{\BBOQ}\APACrefatitle {Verification of Forecasts Expressed in Terms of
  probability} {Verification of forecasts expressed in terms of
  probability}.{\BBCQ}
\newblock
\APACjournalVolNumPages{Monthly Weather Review}{78}{1}{1-3}.
\newblock

\newblock

\PrintBackRefs{\CurrentBib}

\bibitem [\protect \citeauthoryear {%
Br{\"{o}}cker%
\ \BBA {} Smith%
}{%
Br{\"{o}}cker%
\ \BBA {} Smith%
}{%
{\protect \APACyear {2007}}%
}]{%
brocker2007}
\APACinsertmetastar {%
brocker2007}%
\begin{APACrefauthors}%
Br{\"{o}}cker, J.%
\BCBT {}\ \BBA {} Smith, L.A.%
\end{APACrefauthors}%
\unskip\
\newblock
\APACrefYearMonthDay{2007}{}{}.
\newblock
{\BBOQ}\APACrefatitle {{Increasing the Reliability of Reliability Diagrams}}
  {{Increasing the Reliability of Reliability Diagrams}}.{\BBCQ}
\newblock
\APACjournalVolNumPages{Weather and Forecasting}{22}{3}{651--661}.
\newblock

\newblock

\PrintBackRefs{\CurrentBib}

\bibitem [\protect \citeauthoryear {%
Bross%
}{%
Bross%
}{%
{\protect \APACyear {1954}}%
}]{%
bross1953design}
\APACinsertmetastar {%
bross1953design}%
\begin{APACrefauthors}%
Bross, I.%
\end{APACrefauthors}%
\unskip\
\newblock
\APACrefYearMonthDay{1954}{}{}.
\newblock
{\BBOQ}\APACrefatitle {Design for decision} {Design for decision}.{\BBCQ}
\newblock
\APACjournalVolNumPages{Science Education}{38}{4}{325-325}.
\newblock

\newblock

\PrintBackRefs{\CurrentBib}

\bibitem [\protect \citeauthoryear {%
Chung%
, Neiswanger%
, Char%
\BCBL {}\ \BBA {} Schneider%
}{%
Chung%
\ \protect \BOthers {.}}{%
{\protect \APACyear {2021}}%
}]{%
chung2021beyond}
\APACinsertmetastar {%
chung2021beyond}%
\begin{APACrefauthors}%
Chung, Y.%
, Neiswanger, W.%
, Char, I.%
\BCBL {} Schneider, J.%
\end{APACrefauthors}%
\unskip\
\newblock
\APACrefYearMonthDay{2021}{}{}.
\newblock
{\BBOQ}\APACrefatitle {Beyond pinball loss: Quantile methods for calibrated
  uncertainty quantification} {Beyond pinball loss: Quantile methods for
  calibrated uncertainty quantification}.{\BBCQ}
\newblock
\APACjournalVolNumPages{Advances in Neural Information Processing
  Systems}{34}{}{10971--10984}.
\newblock

\newblock

\PrintBackRefs{\CurrentBib}

\bibitem [\protect \citeauthoryear {%
{Cobb}%
, {Roberts}%
\BCBL {}\ \BBA {} {Gal}%
}{%
{Cobb}%
\ \protect \BOthers {.}}{%
{\protect \APACyear {2018}}%
}]{%
CobbGal2018Loss}
\APACinsertmetastar {%
CobbGal2018Loss}%
\begin{APACrefauthors}%
{Cobb}, A.D.%
, {Roberts}, S.J.%
\BCBL {} {Gal}, Y.%
\end{APACrefauthors}%
\unskip\
\newblock
\APACrefYearMonthDay{2018}{}{}.
\newblock
{\BBOQ}\APACrefatitle {{Loss-Calibrated Approximate Inference in Bayesian
  Neural Networks}} {{Loss-Calibrated Approximate Inference in Bayesian Neural
  Networks}}.{\BBCQ}
\newblock
 \APACrefbtitle {Theory of deep learning workshop, {ICML}.} {Theory of deep
  learning workshop, {ICML}.}
\PrintBackRefs{\CurrentBib}

\bibitem [\protect \citeauthoryear {%
Cui%
, Hu%
\BCBL {}\ \BBA {} Zhu%
}{%
Cui%
\ \protect \BOthers {.}}{%
{\protect \APACyear {2020}}%
}]{%
cui2020calibrated}
\APACinsertmetastar {%
cui2020calibrated}%
\begin{APACrefauthors}%
Cui, P.%
, Hu, W.%
\BCBL {} Zhu, J.%
\end{APACrefauthors}%
\unskip\
\newblock
\APACrefYearMonthDay{2020}{}{}.
\newblock
{\BBOQ}\APACrefatitle {Calibrated Reliable Regression using Maximum Mean
  Discrepancy} {Calibrated reliable regression using maximum mean
  discrepancy}.{\BBCQ}
\newblock
\APACjournalVolNumPages{Advances in Neural Information Processing
  Systems}{33}{}{}.
\newblock

\newblock

\PrintBackRefs{\CurrentBib}

\bibitem [\protect \citeauthoryear {%
DeGroot%
\ \BBA {} Fienberg%
}{%
DeGroot%
\ \BBA {} Fienberg%
}{%
{\protect \APACyear {1983}}%
}]{%
degroot1983}
\APACinsertmetastar {%
degroot1983}%
\begin{APACrefauthors}%
DeGroot, M.H.%
\BCBT {}\ \BBA {} Fienberg, S.E.%
\end{APACrefauthors}%
\unskip\
\newblock
\APACrefYearMonthDay{1983}{}{}.
\newblock
{\BBOQ}\APACrefatitle {{The Comparison and Evaluation of Forecasters}} {{The
  Comparison and Evaluation of Forecasters}}.{\BBCQ}
\newblock
\APACjournalVolNumPages{Journal of the Royal Statistical Society. Series D (The
  Statistician)}{32}{1/2}{12--22}.
\newblock

\newblock

\PrintBackRefs{\CurrentBib}

\bibitem [\protect \citeauthoryear {%
Ding%
, Han%
, Liu%
\BCBL {}\ \BBA {} Niethammer%
}{%
Ding%
\ \protect \BOthers {.}}{%
{\protect \APACyear {2021}}%
}]{%
ding2021local}
\APACinsertmetastar {%
ding2021local}%
\begin{APACrefauthors}%
Ding, Z.%
, Han, X.%
, Liu, P.%
\BCBL {} Niethammer, M.%
\end{APACrefauthors}%
\unskip\
\newblock
\APACrefYearMonthDay{2021}{}{}.
\newblock
{\BBOQ}\APACrefatitle {Local temperature scaling for probability calibration}
  {Local temperature scaling for probability calibration}.{\BBCQ}
\newblock
 \APACrefbtitle {Proceedings of the IEEE/CVF International Conference on
  Computer Vision} {Proceedings of the ieee/cvf international conference on
  computer vision}\ (\BPGS\ 6889--6899).
\PrintBackRefs{\CurrentBib}

\bibitem [\protect \citeauthoryear {%
Fagerland%
, Hosmer%
\BCBL {}\ \BBA {} Bofin%
}{%
Fagerland%
\ \protect \BOthers {.}}{%
{\protect \APACyear {2008}}%
}]{%
fagerland2008multinomial}
\APACinsertmetastar {%
fagerland2008multinomial}%
\begin{APACrefauthors}%
Fagerland, M.W.%
, Hosmer, D.W.%
\BCBL {} Bofin, A.M.%
\end{APACrefauthors}%
\unskip\
\newblock
\APACrefYearMonthDay{2008}{}{}.
\newblock
{\BBOQ}\APACrefatitle {Multinomial goodness-of-fit tests for logistic
  regression models} {Multinomial goodness-of-fit tests for logistic regression
  models}.{\BBCQ}
\newblock
\APACjournalVolNumPages{Statistics in Medicine}{27}{21}{4238-4253}.
\newblock
\begin{APACrefURL} {https://onlinelibrary.wiley.com/doi/abs/10.1002/sim.3202}
  \end{APACrefURL}
\newblock
{\href{https://arxiv.org/abs/https://onlinelibrary.wiley.com/doi/pdf/10.1002/sim.3202}{{https://onlinelibrary.wiley.com/doi/pdf/10.1002/sim.3202}}}
\newblock

\newblock
\begin{APACrefDOI} \doi{https://doi.org/10.1002/sim.3202} \end{APACrefDOI}
\PrintBackRefs{\CurrentBib}

\bibitem [\protect \citeauthoryear {%
Fasiolo%
, Wood%
, Zaffran%
, Nedellec%
\BCBL {}\ \BBA {} Goude%
}{%
Fasiolo%
\ \protect \BOthers {.}}{%
{\protect \APACyear {2020}}%
}]{%
fasiolo2020fast}
\APACinsertmetastar {%
fasiolo2020fast}%
\begin{APACrefauthors}%
Fasiolo, M.%
, Wood, S.N.%
, Zaffran, M.%
, Nedellec, R.%
\BCBL {} Goude, Y.%
\end{APACrefauthors}%
\unskip\
\newblock
\APACrefYearMonthDay{2020}{}{}.
\newblock
{\BBOQ}\APACrefatitle {Fast calibrated additive quantile regression} {Fast
  calibrated additive quantile regression}.{\BBCQ}
\newblock
\APACjournalVolNumPages{Journal of the American Statistical
  Association}{}{}{1--11}.
\newblock

\newblock

\PrintBackRefs{\CurrentBib}

\bibitem [\protect \citeauthoryear {%
Fawcett%
\ \BBA {} Niculescu-Mizil%
}{%
Fawcett%
\ \BBA {} Niculescu-Mizil%
}{%
{\protect \APACyear {2007}}%
}]{%
fawcett2007}
\APACinsertmetastar {%
fawcett2007}%
\begin{APACrefauthors}%
Fawcett, T.%
\BCBT {}\ \BBA {} Niculescu-Mizil, A.%
\end{APACrefauthors}%
\unskip\
\newblock
\APACrefYearMonthDay{2007}{}{}.
\newblock
{\BBOQ}\APACrefatitle {{PAV and the ROC convex hull}} {{PAV and the ROC convex
  hull}}.{\BBCQ}
\newblock
\APACjournalVolNumPages{Machine Learning}{68}{1}{97--106}.
\newblock

\newblock

\PrintBackRefs{\CurrentBib}

\bibitem [\protect \citeauthoryear {%
P.~Flach%
}{%
P.~Flach%
}{%
{\protect \APACyear {2012}}%
}]{%
flach2012machine}
\APACinsertmetastar {%
flach2012machine}%
\begin{APACrefauthors}%
Flach, P.%
\end{APACrefauthors}%
\unskip\
\newblock
\APACrefYear{2012}.
\newblock
\APACrefbtitle {Machine learning: the art and science of algorithms that make
  sense of data} {Machine learning: the art and science of algorithms that make
  sense of data}.
\newblock
\APACaddressPublisher{}{Cambridge University Press}.
\PrintBackRefs{\CurrentBib}

\bibitem [\protect \citeauthoryear {%
P.~Flach%
}{%
P.~Flach%
}{%
{\protect \APACyear {2014}}%
}]{%
flach2014classification}
\APACinsertmetastar {%
flach2014classification}%
\begin{APACrefauthors}%
Flach, P.%
\end{APACrefauthors}%
\unskip\
\newblock
\APACrefYearMonthDay{2014}{}{}.
\newblock
{\BBOQ}\APACrefatitle {Classification in context: Adapting to changes in class
  and cost distribution} {Classification in context: Adapting to changes in
  class and cost distribution}.{\BBCQ}
\newblock
 \APACrefbtitle {First international workshop on learning over multiple
  contexts at {European} conference on machine learning and principles and
  practice of knowledge discovery in databases ({ECML-PKDD}).} {First
  international workshop on learning over multiple contexts at {European}
  conference on machine learning and principles and practice of knowledge
  discovery in databases ({ECML-PKDD}).}
\PrintBackRefs{\CurrentBib}

\bibitem [\protect \citeauthoryear {%
P.A.~Flach%
\ \BBA {} Kull%
}{%
P.A.~Flach%
\ \BBA {} Kull%
}{%
{\protect \APACyear {2015}}%
}]{%
flach2015}
\APACinsertmetastar {%
flach2015}%
\begin{APACrefauthors}%
Flach, P.A.%
\BCBT {}\ \BBA {} Kull, M.%
\end{APACrefauthors}%
\unskip\
\newblock
\APACrefYearMonthDay{2015}{}{}.
\newblock
{\BBOQ}\APACrefatitle {{Precision-Recall-Gain Curves: PR Analysis Done Right}}
  {{Precision-Recall-Gain Curves: PR Analysis Done Right}}.{\BBCQ}
\newblock
 \APACrefbtitle {Advances in Neural Information Processing Systems (NIPS'15)}
  {Advances in neural information processing systems (nips'15)}\ (\BPGS\
  838--846).
\PrintBackRefs{\CurrentBib}

\bibitem [\protect \citeauthoryear {%
Ghandeharioun%
, Eoff%
, Jou%
\BCBL {}\ \BBA {} Picard%
}{%
Ghandeharioun%
\ \protect \BOthers {.}}{%
{\protect \APACyear {2019}}%
}]{%
ghandeharioun2019characterizing}
\APACinsertmetastar {%
ghandeharioun2019characterizing}%
\begin{APACrefauthors}%
Ghandeharioun, A.%
, Eoff, B.%
, Jou, B.%
\BCBL {} Picard, R.%
\end{APACrefauthors}%
\unskip\
\newblock
\APACrefYearMonthDay{2019}{}{}.
\newblock
{\BBOQ}\APACrefatitle {Characterizing Sources of Uncertainty to Proxy
  Calibration and Disambiguate Annotator and Data Bias} {Characterizing sources
  of uncertainty to proxy calibration and disambiguate annotator and data
  bias}.{\BBCQ}
\newblock
 \APACrefbtitle {2019 IEEE/CVF International Conference on Computer Vision
  Workshop ({ICCVW})} {2019 ieee/cvf international conference on computer
  vision workshop ({ICCVW})}\ (\BPGS\ 4202--4206).
\PrintBackRefs{\CurrentBib}

\bibitem [\protect \citeauthoryear {%
Gneiting%
\ \BBA {} Raftery%
}{%
Gneiting%
\ \BBA {} Raftery%
}{%
{\protect \APACyear {2007}}%
}]{%
gneiting2007}
\APACinsertmetastar {%
gneiting2007}%
\begin{APACrefauthors}%
Gneiting, T.%
\BCBT {}\ \BBA {} Raftery, A.E.%
\end{APACrefauthors}%
\unskip\
\newblock
\APACrefYearMonthDay{2007}{}{}.
\newblock
{\BBOQ}\APACrefatitle {{Strictly proper scoring rules, prediction, and
  estimation}} {{Strictly proper scoring rules, prediction, and
  estimation}}.{\BBCQ}
\newblock
\APACjournalVolNumPages{Journal of the American Statistical
  Association}{102}{477}{359--378}.
\newblock

\newblock

\PrintBackRefs{\CurrentBib}

\bibitem [\protect \citeauthoryear {%
Gretton%
, Borgwardt%
, Rasch%
, Sch{\"o}lkopf%
\BCBL {}\ \BBA {} Smola%
}{%
Gretton%
\ \protect \BOthers {.}}{%
{\protect \APACyear {2012}}%
}]{%
gretton2012kernel}
\APACinsertmetastar {%
gretton2012kernel}%
\begin{APACrefauthors}%
Gretton, A.%
, Borgwardt, K.M.%
, Rasch, M.J.%
, Sch{\"o}lkopf, B.%
\BCBL {} Smola, A.%
\end{APACrefauthors}%
\unskip\
\newblock
\APACrefYearMonthDay{2012}{}{}.
\newblock
{\BBOQ}\APACrefatitle {A kernel two-sample test} {A kernel two-sample
  test}.{\BBCQ}
\newblock
\APACjournalVolNumPages{The Journal of Machine Learning
  Research}{13}{1}{723--773}.
\newblock

\newblock

\PrintBackRefs{\CurrentBib}

\bibitem [\protect \citeauthoryear {%
Guillory%
, Shankar%
, Ebrahimi%
, Darrell%
\BCBL {}\ \BBA {} Schmidt%
}{%
Guillory%
\ \protect \BOthers {.}}{%
{\protect \APACyear {2021}}%
}]{%
guillory2021predicting}
\APACinsertmetastar {%
guillory2021predicting}%
\begin{APACrefauthors}%
Guillory, D.%
, Shankar, V.%
, Ebrahimi, S.%
, Darrell, T.%
\BCBL {} Schmidt, L.%
\end{APACrefauthors}%
\unskip\
\newblock
\APACrefYearMonthDay{2021}{}{}.
\newblock
{\BBOQ}\APACrefatitle {Predicting with confidence on unseen distributions}
  {Predicting with confidence on unseen distributions}.{\BBCQ}
\newblock
 \APACrefbtitle {Proceedings of the IEEE/CVF International Conference on
  Computer Vision} {Proceedings of the ieee/cvf international conference on
  computer vision}\ (\BPGS\ 1134--1144).
\PrintBackRefs{\CurrentBib}

\bibitem [\protect \citeauthoryear {%
Guo%
, Pleiss%
, Sun%
\BCBL {}\ \BBA {} Weinberger%
}{%
Guo%
\ \protect \BOthers {.}}{%
{\protect \APACyear {2017}}%
}]{%
guo2017}
\APACinsertmetastar {%
guo2017}%
\begin{APACrefauthors}%
Guo, C.%
, Pleiss, G.%
, Sun, Y.%
\BCBL {} Weinberger, K.Q.%
\end{APACrefauthors}%
\unskip\
\newblock
\APACrefYearMonthDay{2017}{}{}.
\newblock
{\BBOQ}\APACrefatitle {{On Calibration of Modern Neural Networks}} {{On
  Calibration of Modern Neural Networks}}.{\BBCQ}
\newblock
 \APACrefbtitle {34th International Conference on Machine Learning} {34th
  international conference on machine learning}\ (\BPGS\ 1321--1330).
\newblock
\APACaddressPublisher{Sydney, Australia}{}.
\PrintBackRefs{\CurrentBib}

\bibitem [\protect \citeauthoryear {%
Hagedorn%
, Doblas-Reyes%
\BCBL {}\ \BBA {} Palmer%
}{%
Hagedorn%
\ \protect \BOthers {.}}{%
{\protect \APACyear {2005}}%
}]{%
hagedorn2005}
\APACinsertmetastar {%
hagedorn2005}%
\begin{APACrefauthors}%
Hagedorn, R.%
, Doblas-Reyes, F.J.%
\BCBL {} Palmer, T.N.%
\end{APACrefauthors}%
\unskip\
\newblock
\APACrefYearMonthDay{2005}{}{}.
\newblock
{\BBOQ}\APACrefatitle {{The rationale behind the success of multi-model
  ensembles in seasonal forecasting — I. Basic concept}} {{The rationale
  behind the success of multi-model ensembles in seasonal forecasting — I.
  Basic concept}}.{\BBCQ}
\newblock
\APACjournalVolNumPages{Tellus A: Dynamic Meteorology and
  Oceanography}{57}{3}{219--233}.
\newblock

\newblock

\PrintBackRefs{\CurrentBib}

\bibitem [\protect \citeauthoryear {%
Hallenbeck%
}{%
Hallenbeck%
}{%
{\protect \APACyear {1920}}%
}]{%
hallenbeck1920}
\APACinsertmetastar {%
hallenbeck1920}%
\begin{APACrefauthors}%
Hallenbeck, C.%
\end{APACrefauthors}%
\unskip\
\newblock
\APACrefYearMonthDay{1920}{}{}.
\newblock
{\BBOQ}\APACrefatitle {{Forecasting Precipitation in Percentages of
  Probability.}} {{Forecasting Precipitation in Percentages of
  Probability.}}{\BBCQ}
\newblock
\APACjournalVolNumPages{Monthly Weather Review}{48}{11}{645--647}.
\newblock

\newblock

\PrintBackRefs{\CurrentBib}

\bibitem [\protect \citeauthoryear {%
Hinton%
, Vinyals%
, Dean%
\BCBL {}\ \protect \BOthers {.}}{%
Hinton%
\ \protect \BOthers {.}}{%
{\protect \APACyear {2015}}%
}]{%
hinton2015distilling}
\APACinsertmetastar {%
hinton2015distilling}%
\begin{APACrefauthors}%
Hinton, G.%
, Vinyals, O.%
, Dean, J.%
\BCBL {}\ \BOthersPeriod {.}\end{APACrefauthors}%
\unskip\
\newblock
\APACrefYearMonthDay{2015}{}{}.
\newblock
{\BBOQ}\APACrefatitle {Distilling the knowledge in a neural network}
  {Distilling the knowledge in a neural network}.{\BBCQ}
\newblock
\APACjournalVolNumPages{arXiv preprint arXiv:1503.02531}{2}{7}{}.
\newblock

\newblock

\PrintBackRefs{\CurrentBib}

\bibitem [\protect \citeauthoryear {%
Hosmer%
\ \BBA {} Lemeshow%
}{%
Hosmer%
\ \BBA {} Lemeshow%
}{%
{\protect \APACyear {2003}}%
}]{%
Hosmer2003applied}
\APACinsertmetastar {%
Hosmer2003applied}%
\begin{APACrefauthors}%
Hosmer, D.%
\BCBT {}\ \BBA {} Lemeshow, S.%
\end{APACrefauthors}%
\unskip\
\newblock
\APACrefYear{2003}.
\newblock
\APACrefbtitle {Applied Logistic Regression} {Applied logistic regression}.
\newblock
\APACaddressPublisher{}{Wiley}.
\PrintBackRefs{\CurrentBib}

\bibitem [\protect \citeauthoryear {%
H\"ullermeier%
, Destercke%
\BCBL {}\ \BBA {} Shaker%
}{%
H\"ullermeier%
\ \protect \BOthers {.}}{%
{\protect \APACyear {2022}}%
}]{%
hullermeier2022quantifying}
\APACinsertmetastar {%
hullermeier2022quantifying}%
\begin{APACrefauthors}%
H\"ullermeier, E.%
, Destercke, S.%
\BCBL {} Shaker, M.H.%
\end{APACrefauthors}%
\unskip\
\newblock
\APACrefYearMonthDay{2022}{01--05 Aug}{}.
\newblock
{\BBOQ}\APACrefatitle {Quantification of Credal Uncertainty in Machine
  Learning: A Critical Analysis and Empirical Comparison} {Quantification of
  credal uncertainty in machine learning: A critical analysis and empirical
  comparison}.{\BBCQ}
\newblock
 J.~Cussens\ \BBA {} K.~Zhang\ (\BEDS), \APACrefbtitle {Proceedings of the
  Thirty-Eighth Conference on Uncertainty in Artificial Intelligence}
  {Proceedings of the thirty-eighth conference on uncertainty in artificial
  intelligence}\ (\BVOL~180, \BPGS\ 548--557).
\newblock
\APACaddressPublisher{}{PMLR}.
\newblock
\begin{APACrefURL} {https://proceedings.mlr.press/v180/hullermeier22a.html}
  \end{APACrefURL}
\PrintBackRefs{\CurrentBib}

\bibitem [\protect \citeauthoryear {%
H{\"u}llermeier%
\ \BBA {} Waegeman%
}{%
H{\"u}llermeier%
\ \BBA {} Waegeman%
}{%
{\protect \APACyear {2021}}%
}]{%
hullermeier2021aleatoric}
\APACinsertmetastar {%
hullermeier2021aleatoric}%
\begin{APACrefauthors}%
H{\"u}llermeier, E.%
\BCBT {}\ \BBA {} Waegeman, W.%
\end{APACrefauthors}%
\unskip\
\newblock
\APACrefYearMonthDay{2021}{}{}.
\newblock
{\BBOQ}\APACrefatitle {Aleatoric and epistemic uncertainty in machine learning:
  An introduction to concepts and methods} {Aleatoric and epistemic uncertainty
  in machine learning: An introduction to concepts and methods}.{\BBCQ}
\newblock
\APACjournalVolNumPages{Machine Learning}{110}{3}{457--506}.
\newblock

\newblock

\PrintBackRefs{\CurrentBib}

\bibitem [\protect \citeauthoryear {%
Jiang%
, Nagarajan%
, Baek%
\BCBL {}\ \BBA {} Kolter%
}{%
Jiang%
\ \protect \BOthers {.}}{%
{\protect \APACyear {2021}}%
}]{%
jiang2021assessing}
\APACinsertmetastar {%
jiang2021assessing}%
\begin{APACrefauthors}%
Jiang, Y.%
, Nagarajan, V.%
, Baek, C.%
\BCBL {} Kolter, J.Z.%
\end{APACrefauthors}%
\unskip\
\newblock
\APACrefYearMonthDay{2021}{}{}.
\newblock
{\BBOQ}\APACrefatitle {Assessing generalization of sgd via disagreement}
  {Assessing generalization of sgd via disagreement}.{\BBCQ}
\newblock
\APACjournalVolNumPages{arXiv preprint arXiv:2106.13799}{}{}{}.
\newblock

\newblock

\PrintBackRefs{\CurrentBib}

\bibitem [\protect \citeauthoryear {%
Johansson%
\ \protect \BOthers {.}}{%
Johansson%
\ \protect \BOthers {.}}{%
{\protect \APACyear {2018}}%
}]{%
johansson2018}
\APACinsertmetastar {%
johansson2018}%
\begin{APACrefauthors}%
Johansson, U.%
, L{\"{o}}fstr{\"{o}}m, T.%
, Sundell, H.%
, Linusson, H.%
, Gidenstam, A.%
\BCBL {} Bostr{\"{o}}m, H.%
\end{APACrefauthors}%
\unskip\
\newblock
\APACrefYearMonthDay{2018}{}{}.
\newblock
{\BBOQ}\APACrefatitle {{Venn predictors for well-calibrated probability
  estimation trees}} {{Venn predictors for well-calibrated probability
  estimation trees}}.{\BBCQ}
\newblock
 A.~Gammerman, V.~Vovk, Z.~Luo, E.~Smirnov\BCBL {}\ \BBA {} R.~Peeters\
  (\BEDS), \APACrefbtitle {Proceedings of the Seventh Workshop on Conformal and
  Probabilistic Prediction and Applications} {Proceedings of the seventh
  workshop on conformal and probabilistic prediction and applications}\ (\BPGS\
  3--14).
\newblock
\APACaddressPublisher{}{PMLR}.
\PrintBackRefs{\CurrentBib}

\bibitem [\protect \citeauthoryear {%
Koenker%
\ \BBA {} Hallock%
}{%
Koenker%
\ \BBA {} Hallock%
}{%
{\protect \APACyear {2001}}%
}]{%
koenker2001quantile}
\APACinsertmetastar {%
koenker2001quantile}%
\begin{APACrefauthors}%
Koenker, R.%
\BCBT {}\ \BBA {} Hallock, K.F.%
\end{APACrefauthors}%
\unskip\
\newblock
\APACrefYearMonthDay{2001}{}{}.
\newblock
{\BBOQ}\APACrefatitle {Quantile regression} {Quantile regression}.{\BBCQ}
\newblock
\APACjournalVolNumPages{Journal of economic perspectives}{15}{4}{143--156}.
\newblock

\newblock

\PrintBackRefs{\CurrentBib}

\bibitem [\protect \citeauthoryear {%
Kristiadi%
, Hein%
\BCBL {}\ \BBA {} Hennig%
}{%
Kristiadi%
\ \protect \BOthers {.}}{%
{\protect \APACyear {2020}}%
}]{%
kristiadi2020being}
\APACinsertmetastar {%
kristiadi2020being}%
\begin{APACrefauthors}%
Kristiadi, A.%
, Hein, M.%
\BCBL {} Hennig, P.%
\end{APACrefauthors}%
\unskip\
\newblock
\APACrefYearMonthDay{2020}{}{}.
\newblock
{\BBOQ}\APACrefatitle {Being bayesian, even just a bit, fixes overconfidence in
  relu networks} {Being bayesian, even just a bit, fixes overconfidence in relu
  networks}.{\BBCQ}
\newblock
 \APACrefbtitle {International Conference on Machine Learning} {International
  conference on machine learning}\ (\BPGS\ 5436--5446).
\PrintBackRefs{\CurrentBib}

\bibitem [\protect \citeauthoryear {%
Kuleshov%
\ \BBA {} Deshpande%
}{%
Kuleshov%
\ \BBA {} Deshpande%
}{%
{\protect \APACyear {2022}}%
}]{%
kuleshov2022calibrated}
\APACinsertmetastar {%
kuleshov2022calibrated}%
\begin{APACrefauthors}%
Kuleshov, V.%
\BCBT {}\ \BBA {} Deshpande, S.%
\end{APACrefauthors}%
\unskip\
\newblock
\APACrefYearMonthDay{2022}{}{}.
\newblock
{\BBOQ}\APACrefatitle {Calibrated and sharp uncertainties in deep learning via
  density estimation} {Calibrated and sharp uncertainties in deep learning via
  density estimation}.{\BBCQ}
\newblock
 \APACrefbtitle {International Conference on Machine Learning} {International
  conference on machine learning}\ (\BPGS\ 11683--11693).
\PrintBackRefs{\CurrentBib}

\bibitem [\protect \citeauthoryear {%
Kuleshov%
, Fenner%
\BCBL {}\ \BBA {} Ermon%
}{%
Kuleshov%
\ \protect \BOthers {.}}{%
{\protect \APACyear {2018}}%
}]{%
kuleshov2018}
\APACinsertmetastar {%
kuleshov2018}%
\begin{APACrefauthors}%
Kuleshov, V.%
, Fenner, N.%
\BCBL {} Ermon, S.%
\end{APACrefauthors}%
\unskip\
\newblock
\APACrefYearMonthDay{2018}{}{}.
\newblock
{\BBOQ}\APACrefatitle {{Accurate uncertainties for deep learning using
  calibrated regression}} {{Accurate uncertainties for deep learning using
  calibrated regression}}.{\BBCQ}
\newblock
 \APACrefbtitle {35th International Conference on Machine Learning, {ICML}
  2018} {35th international conference on machine learning, {ICML} 2018}\
  (\BVOL~6, \BPGS\ 4369--4377).
\newblock
\APACaddressPublisher{}{International Machine Learning Society (IMLS)}.
\PrintBackRefs{\CurrentBib}

\bibitem [\protect \citeauthoryear {%
Kuleshov%
\ \BBA {} Liang%
}{%
Kuleshov%
\ \BBA {} Liang%
}{%
{\protect \APACyear {2015}}%
}]{%
kuleshov2015calibrated}
\APACinsertmetastar {%
kuleshov2015calibrated}%
\begin{APACrefauthors}%
Kuleshov, V.%
\BCBT {}\ \BBA {} Liang, P.S.%
\end{APACrefauthors}%
\unskip\
\newblock
\APACrefYearMonthDay{2015}{}{}.
\newblock
{\BBOQ}\APACrefatitle {Calibrated structured prediction} {Calibrated structured
  prediction}.{\BBCQ}
\newblock
\APACjournalVolNumPages{Advances in Neural Information Processing
  Systems}{28}{}{3474--3482}.
\newblock

\newblock

\PrintBackRefs{\CurrentBib}

\bibitem [\protect \citeauthoryear {%
Kull%
\ \BBA {} Flach%
}{%
Kull%
\ \BBA {} Flach%
}{%
{\protect \APACyear {2015}}%
}]{%
kull2015novel}
\APACinsertmetastar {%
kull2015novel}%
\begin{APACrefauthors}%
Kull, M.%
\BCBT {}\ \BBA {} Flach, P.%
\end{APACrefauthors}%
\unskip\
\newblock
\APACrefYearMonthDay{2015}{}{}.
\newblock
{\BBOQ}\APACrefatitle {Novel decompositions of proper scoring rules for
  classification: Score adjustment as precursor to calibration} {Novel
  decompositions of proper scoring rules for classification: Score adjustment
  as precursor to calibration}.{\BBCQ}
\newblock
 \APACrefbtitle {Joint European Conference on Machine Learning and Knowledge
  Discovery in Databases} {Joint european conference on machine learning and
  knowledge discovery in databases}\ (\BPGS\ 68--85).
\PrintBackRefs{\CurrentBib}

\bibitem [\protect \citeauthoryear {%
Kull%
\ \protect \BOthers {.}}{%
Kull%
\ \protect \BOthers {.}}{%
{\protect \APACyear {2019}}%
}]{%
kull2019}
\APACinsertmetastar {%
kull2019}%
\begin{APACrefauthors}%
Kull, M.%
, Perello-Nieto, M.%
, K{\"{a}}ngsepp, M.%
, Filho, T.S.%
, Song, H.%
\BCBL {} Flach, P.%
\end{APACrefauthors}%
\unskip\
\newblock
\APACrefYearMonthDay{2019}{}{}.
\newblock
{\BBOQ}\APACrefatitle {{Beyond temperature scaling: Obtaining well-calibrated
  multiclass probabilities with Dirichlet calibration}} {{Beyond temperature
  scaling: Obtaining well-calibrated multiclass probabilities with Dirichlet
  calibration}}.{\BBCQ}
\newblock
 \APACrefbtitle {Advances in Neural Information Processing Systems
  ({NeurIPS'19})} {Advances in neural information processing systems
  ({NeurIPS'19})}\ (\BPGS\ 12316--12326).
\PrintBackRefs{\CurrentBib}

\bibitem [\protect \citeauthoryear {%
Kull%
, {Silva Filho}%
\BCBL {}\ \BBA {} Flach%
}{%
Kull%
, {Silva Filho}%
\BCBL {}\ \BBA {} Flach%
}{%
{\protect \APACyear {2017}}%
}]{%
pmlr-v54-kull17a}
\APACinsertmetastar {%
pmlr-v54-kull17a}%
\begin{APACrefauthors}%
Kull, M.%
, {Silva Filho}, T.%
\BCBL {} Flach, P.%
\end{APACrefauthors}%
\unskip\
\newblock
\APACrefYearMonthDay{2017}{20--22 Apr}{}.
\newblock
{\BBOQ}\APACrefatitle {{Beta calibration: a well-founded and easily implemented
  improvement on logistic calibration for binary classifiers}} {{Beta
  calibration: a well-founded and easily implemented improvement on logistic
  calibration for binary classifiers}}.{\BBCQ}
\newblock
 A.~Singh\ \BBA {} J.~Zhu\ (\BEDS), \APACrefbtitle {Proceedings of the 20th
  International Conference on Artificial Intelligence and Statistics}
  {Proceedings of the 20th international conference on artificial intelligence
  and statistics}\ (\BVOL~54, \BPGS\ 623--631).
\newblock
\APACaddressPublisher{Fort Lauderdale, FL, USA}{PMLR}.
\PrintBackRefs{\CurrentBib}

\bibitem [\protect \citeauthoryear {%
Kull%
, Silva~Filho%
\BCBL {}\ \BBA {} Flach%
}{%
Kull%
, Silva~Filho%
\BCBL {}\ \BBA {} Flach%
}{%
{\protect \APACyear {2017}}%
}]{%
kull2017beyond}
\APACinsertmetastar {%
kull2017beyond}%
\begin{APACrefauthors}%
Kull, M.%
, Silva~Filho, T.M.%
\BCBL {} Flach, P.%
\end{APACrefauthors}%
\unskip\
\newblock
\APACrefYearMonthDay{2017}{}{}.
\newblock
{\BBOQ}\APACrefatitle {Beyond sigmoids: How to obtain well-calibrated
  probabilities from binary classifiers with beta calibration} {Beyond
  sigmoids: How to obtain well-calibrated probabilities from binary classifiers
  with beta calibration}.{\BBCQ}
\newblock
\APACjournalVolNumPages{Electronic Journal of Statistics}{11}{2}{5052--5080}.
\newblock

\newblock

\PrintBackRefs{\CurrentBib}

\bibitem [\protect \citeauthoryear {%
Kumar%
, Liang%
\BCBL {}\ \BBA {} Ma%
}{%
Kumar%
\ \protect \BOthers {.}}{%
{\protect \APACyear {2019}}%
}]{%
kumar2019}
\APACinsertmetastar {%
kumar2019}%
\begin{APACrefauthors}%
Kumar, A.%
, Liang, P.%
\BCBL {} Ma, T.%
\end{APACrefauthors}%
\unskip\
\newblock
\APACrefYearMonthDay{2019}{}{}.
\newblock
{\BBOQ}\APACrefatitle {{Verified Uncertainty Calibration}} {{Verified
  Uncertainty Calibration}}.{\BBCQ}
\newblock
 \APACrefbtitle {Advances in Neural Information Processing Systems
  ({NeurIPS'19}).} {Advances in neural information processing systems
  ({NeurIPS'19}).}
\PrintBackRefs{\CurrentBib}

\bibitem [\protect \citeauthoryear {%
Kumar%
, Sarawagi%
\BCBL {}\ \BBA {} Jain%
}{%
Kumar%
\ \protect \BOthers {.}}{%
{\protect \APACyear {2018}}%
}]{%
kumar2018}
\APACinsertmetastar {%
kumar2018}%
\begin{APACrefauthors}%
Kumar, A.%
, Sarawagi, S.%
\BCBL {} Jain, U.%
\end{APACrefauthors}%
\unskip\
\newblock
\APACrefYearMonthDay{2018}{10--15 Jul}{}.
\newblock
{\BBOQ}\APACrefatitle {{Trainable Calibration Measures For Neural Networks From
  Kernel Mean Embeddings}} {{Trainable Calibration Measures For Neural Networks
  From Kernel Mean Embeddings}}.{\BBCQ}
\newblock
 J.~Dy\ \BBA {} A.~Krause\ (\BEDS), \APACrefbtitle {35th International
  Conference on Machine Learning, volume 80 of Proceedings of Machine Learning
  Research} {35th international conference on machine learning, volume 80 of
  proceedings of machine learning research}\ (\BVOL~80, \BPGS\ 2805--2814).
\newblock
\APACaddressPublisher{Stockholmsmässan, Stockholm Sweden}{PMLR}.
\PrintBackRefs{\CurrentBib}

\bibitem [\protect \citeauthoryear {%
Kängsepp%
, Valk%
\BCBL {}\ \BBA {} Kull%
}{%
Kängsepp%
\ \protect \BOthers {.}}{%
{\protect \APACyear {2022}}%
}]{%
kangsepp2022usefulness}
\APACinsertmetastar {%
kangsepp2022usefulness}%
\begin{APACrefauthors}%
Kängsepp, M.%
, Valk, K.%
\BCBL {} Kull, M.%
\end{APACrefauthors}%
\unskip\
\newblock
\APACrefYearMonthDay{2022}{}{}.
\newblock
\APACrefbtitle {On the Usefulness of the Fit-on-the-Test View on Evaluating
  Calibration of Classifiers.} {On the usefulness of the fit-on-the-test view
  on evaluating calibration of classifiers.}
\newblock
\APACaddressPublisher{}{arXiv}.
\newblock
\begin{APACrefURL} {https://arxiv.org/abs/2203.08958} \end{APACrefURL}
\newblock
\begin{APACrefDOI} \doi{10.48550/ARXIV.2203.08958} \end{APACrefDOI}
\PrintBackRefs{\CurrentBib}

\bibitem [\protect \citeauthoryear {%
Lachiche%
\ \BBA {} Flach%
}{%
Lachiche%
\ \BBA {} Flach%
}{%
{\protect \APACyear {2003}}%
}]{%
lachiche2003improving}
\APACinsertmetastar {%
lachiche2003improving}%
\begin{APACrefauthors}%
Lachiche, N.%
\BCBT {}\ \BBA {} Flach, P.A.%
\end{APACrefauthors}%
\unskip\
\newblock
\APACrefYearMonthDay{2003}{}{}.
\newblock
{\BBOQ}\APACrefatitle {Improving accuracy and cost of two-class and multi-class
  probabilistic classifiers using ROC curves} {Improving accuracy and cost of
  two-class and multi-class probabilistic classifiers using roc curves}.{\BBCQ}
\newblock
 \APACrefbtitle {Proceedings of the 20th International Conference on Machine
  Learning ({ICML-03})} {Proceedings of the 20th international conference on
  machine learning ({ICML-03})}\ (\BPGS\ 416--423).
\PrintBackRefs{\CurrentBib}

\bibitem [\protect \citeauthoryear {%
Leathart%
, Frank%
, Holmes%
\BCBL {}\ \BBA {} Pfahringer%
}{%
Leathart%
\ \protect \BOthers {.}}{%
{\protect \APACyear {2017}}%
}]{%
leathart2017}
\APACinsertmetastar {%
leathart2017}%
\begin{APACrefauthors}%
Leathart, T.%
, Frank, E.%
, Holmes, G.%
\BCBL {} Pfahringer, B.%
\end{APACrefauthors}%
\unskip\
\newblock
\APACrefYearMonthDay{2017}{jul}{}.
\newblock
{\BBOQ}\APACrefatitle {{Probability Calibration Trees}} {{Probability
  Calibration Trees}}.{\BBCQ}
\newblock
 \APACrefbtitle {Asian Conference on Machine Learning} {Asian conference on
  machine learning}\ (\BPGS\ 145--160).
\PrintBackRefs{\CurrentBib}

\bibitem [\protect \citeauthoryear {%
Lichtenstein%
, Fischhoff%
\BCBL {}\ \BBA {} Phillips%
}{%
Lichtenstein%
\ \protect \BOthers {.}}{%
{\protect \APACyear {1977}}%
}]{%
lichtenstein1977calibration}
\APACinsertmetastar {%
lichtenstein1977calibration}%
\begin{APACrefauthors}%
Lichtenstein, S.%
, Fischhoff, B.%
\BCBL {} Phillips, L.D.%
\end{APACrefauthors}%
\unskip\
\newblock
\APACrefYearMonthDay{1977}{}{}.
\newblock
{\BBOQ}\APACrefatitle {Calibration of probabilities: The state of the art}
  {Calibration of probabilities: The state of the art}.{\BBCQ}
\newblock
 \APACrefbtitle {Decision making and change in human affairs} {Decision making
  and change in human affairs}\ (\BPGS\ 275--324).
\newblock
\APACaddressPublisher{}{Springer}.
\PrintBackRefs{\CurrentBib}

\bibitem [\protect \citeauthoryear {%
Lichtenstein%
, Fischhoff%
\BCBL {}\ \BBA {} Phillips%
}{%
Lichtenstein%
\ \protect \BOthers {.}}{%
{\protect \APACyear {1982}}%
}]{%
lichtenstein1982}
\APACinsertmetastar {%
lichtenstein1982}%
\begin{APACrefauthors}%
Lichtenstein, S.%
, Fischhoff, B.%
\BCBL {} Phillips, L.D.%
\end{APACrefauthors}%
\unskip\
\newblock
\APACrefYearMonthDay{1982}{}{}.
\newblock
{\BBOQ}\APACrefatitle {{Calibration of probabilities: The state of the art to
  1980}} {{Calibration of probabilities: The state of the art to 1980}}.{\BBCQ}
\newblock
\BIn{} \APACrefbtitle {Judgment under Uncertainty: Heuristics and Biases}
  {Judgment under uncertainty: Heuristics and biases}\ (\BPGS\ 306--334).
\newblock
\APACaddressPublisher{}{Cambridge University Press}.
\PrintBackRefs{\CurrentBib}

\bibitem [\protect \citeauthoryear {%
J.Z.~Liu%
, Paisley%
, Kioumourtzoglou%
\BCBL {}\ \BBA {} Coull%
}{%
J.Z.~Liu%
\ \protect \BOthers {.}}{%
{\protect \APACyear {2019}}%
}]{%
liu2019accurate}
\APACinsertmetastar {%
liu2019accurate}%
\begin{APACrefauthors}%
Liu, J.Z.%
, Paisley, J.%
, Kioumourtzoglou, M\BHBI A.%
\BCBL {} Coull, B.%
\end{APACrefauthors}%
\unskip\
\newblock
\APACrefYearMonthDay{2019}{}{}.
\newblock
{\BBOQ}\APACrefatitle {Accurate uncertainty estimation and decomposition in
  ensemble learning} {Accurate uncertainty estimation and decomposition in
  ensemble learning}.{\BBCQ}
\newblock
\APACjournalVolNumPages{arXiv preprint arXiv:1911.04061}{}{}{}.
\newblock

\newblock

\PrintBackRefs{\CurrentBib}

\bibitem [\protect \citeauthoryear {%
S.~Liu%
, Long%
, Wang%
\BCBL {}\ \BBA {} Jordan%
}{%
S.~Liu%
\ \protect \BOthers {.}}{%
{\protect \APACyear {2018}}%
}]{%
liu2018generalized}
\APACinsertmetastar {%
liu2018generalized}%
\begin{APACrefauthors}%
Liu, S.%
, Long, M.%
, Wang, J.%
\BCBL {} Jordan, M.I.%
\end{APACrefauthors}%
\unskip\
\newblock
\APACrefYearMonthDay{2018}{}{}.
\newblock
{\BBOQ}\APACrefatitle {Generalized zero-shot learning with deep calibration
  network} {Generalized zero-shot learning with deep calibration
  network}.{\BBCQ}
\newblock
 \APACrefbtitle {Advances in Neural Information Processing Systems} {Advances
  in neural information processing systems}\ (\BPGS\ 2005--2015).
\PrintBackRefs{\CurrentBib}

\bibitem [\protect \citeauthoryear {%
Ma%
\ \BBA {} Blaschko%
}{%
Ma%
\ \BBA {} Blaschko%
}{%
{\protect \APACyear {2021}}%
}]{%
ma2021meta}
\APACinsertmetastar {%
ma2021meta}%
\begin{APACrefauthors}%
Ma, X.%
\BCBT {}\ \BBA {} Blaschko, M.B.%
\end{APACrefauthors}%
\unskip\
\newblock
\APACrefYearMonthDay{2021}{}{}.
\newblock
{\BBOQ}\APACrefatitle {Meta-cal: Well-controlled post-hoc calibration by
  ranking} {Meta-cal: Well-controlled post-hoc calibration by ranking}.{\BBCQ}
\newblock
 \APACrefbtitle {International Conference on Machine Learning} {International
  conference on machine learning}\ (\BPGS\ 7235--7245).
\PrintBackRefs{\CurrentBib}

\bibitem [\protect \citeauthoryear {%
Menon%
\ \BBA {} Williamson%
}{%
Menon%
\ \BBA {} Williamson%
}{%
{\protect \APACyear {2018}}%
}]{%
menon2018loss}
\APACinsertmetastar {%
menon2018loss}%
\begin{APACrefauthors}%
Menon, A.K.%
\BCBT {}\ \BBA {} Williamson, R.C.%
\end{APACrefauthors}%
\unskip\
\newblock
\APACrefYearMonthDay{2018}{}{}.
\newblock
{\BBOQ}\APACrefatitle {A loss framework for calibrated anomaly detection} {A
  loss framework for calibrated anomaly detection}.{\BBCQ}
\newblock
 \APACrefbtitle {Proceedings of the 32nd International Conference on Neural
  Information Processing Systems} {Proceedings of the 32nd international
  conference on neural information processing systems}\ (\BPGS\ 1494--1504).
\PrintBackRefs{\CurrentBib}

\bibitem [\protect \citeauthoryear {%
Milios%
, Michiardi%
, Rosasco%
\BCBL {}\ \BBA {} Filippone%
}{%
Milios%
\ \protect \BOthers {.}}{%
{\protect \APACyear {2018}}%
}]{%
milios2018}
\APACinsertmetastar {%
milios2018}%
\begin{APACrefauthors}%
Milios, D.%
, Michiardi, P.%
, Rosasco, L.%
\BCBL {} Filippone, M.%
\end{APACrefauthors}%
\unskip\
\newblock
\APACrefYearMonthDay{2018}{}{}.
\newblock
{\BBOQ}\APACrefatitle {{Dirichlet-based Gaussian Processes for Large-scale
  Calibrated Classification}} {{Dirichlet-based Gaussian Processes for
  Large-scale Calibrated Classification}}.{\BBCQ}
\newblock
 \APACrefbtitle {Advances in Neural Information Processing Systems ({NIPS'18})}
  {Advances in neural information processing systems ({NIPS'18})}\ (\BPGS\
  6005--6015).
\PrintBackRefs{\CurrentBib}

\bibitem [\protect \citeauthoryear {%
Miller%
}{%
Miller%
}{%
{\protect \APACyear {1962}}%
}]{%
miller1962statistical}
\APACinsertmetastar {%
miller1962statistical}%
\begin{APACrefauthors}%
Miller, R.G.%
\end{APACrefauthors}%
\unskip\
\newblock
\APACrefYearMonthDay{1962}{}{}.
\newblock
{\BBOQ}\APACrefatitle {Statistical Prediction by Discriminant Analysis}
  {Statistical prediction by discriminant analysis}.{\BBCQ}
\newblock
\BIn{} \APACrefbtitle {Statistical Prediction by Discriminant Analysis}
  {Statistical prediction by discriminant analysis}\ (\BPGS\ 1--54).
\newblock
\APACaddressPublisher{Boston, MA}{American Meteorological Society}.
\PrintBackRefs{\CurrentBib}

\bibitem [\protect \citeauthoryear {%
M{\"u}ller%
, Kornblith%
\BCBL {}\ \BBA {} Hinton%
}{%
M{\"u}ller%
\ \protect \BOthers {.}}{%
{\protect \APACyear {2019}}%
}]{%
muller2019does}
\APACinsertmetastar {%
muller2019does}%
\begin{APACrefauthors}%
M{\"u}ller, R.%
, Kornblith, S.%
\BCBL {} Hinton, G.%
\end{APACrefauthors}%
\unskip\
\newblock
\APACrefYearMonthDay{2019}{}{}.
\newblock
{\BBOQ}\APACrefatitle {When does label smoothing help?} {When does label
  smoothing help?}{\BBCQ}
\newblock
\APACjournalVolNumPages{arXiv preprint arXiv:1906.02629}{}{}{}.
\newblock

\newblock

\PrintBackRefs{\CurrentBib}

\bibitem [\protect \citeauthoryear {%
Murphy%
\ \BBA {} Winkler%
}{%
Murphy%
\ \BBA {} Winkler%
}{%
{\protect \APACyear {1977}}%
}]{%
murphy1977reliability}
\APACinsertmetastar {%
murphy1977reliability}%
\begin{APACrefauthors}%
Murphy, A.H.%
\BCBT {}\ \BBA {} Winkler, R.L.%
\end{APACrefauthors}%
\unskip\
\newblock
\APACrefYearMonthDay{1977}{}{}.
\newblock
{\BBOQ}\APACrefatitle {Reliability of Subjective Probability Forecasts of
  Precipitation and Temperature} {Reliability of subjective probability
  forecasts of precipitation and temperature}.{\BBCQ}
\newblock
\APACjournalVolNumPages{Journal of the Royal Statistical Society. Series C
  (Applied Statistics)}{26}{1}{41--47}.
\newblock

\newblock

\PrintBackRefs{\CurrentBib}

\bibitem [\protect \citeauthoryear {%
M.P.~Naeini%
\ \BBA {} Cooper%
}{%
M.P.~Naeini%
\ \BBA {} Cooper%
}{%
{\protect \APACyear {2016}}%
}]{%
naeini2016}
\APACinsertmetastar {%
naeini2016}%
\begin{APACrefauthors}%
Naeini, M.P.%
\BCBT {}\ \BBA {} Cooper, G.F.%
\end{APACrefauthors}%
\unskip\
\newblock
\APACrefYearMonthDay{2016}{}{}.
\newblock
{\BBOQ}\APACrefatitle {{Binary Classifier Calibration Using an Ensemble of Near
  Isotonic Regression Models}} {{Binary Classifier Calibration Using an
  Ensemble of Near Isotonic Regression Models}}.{\BBCQ}
\newblock
 \APACrefbtitle {{IEEE} 16th International Conference on Data Mining ({ICDM})}
  {{IEEE} 16th international conference on data mining ({ICDM})}\ (\BPGS\
  360--369).
\newblock
\APACaddressPublisher{}{Institute of Electrical and Electronics Engineers
  (IEEE)}.
\PrintBackRefs{\CurrentBib}

\bibitem [\protect \citeauthoryear {%
P.~Naeini%
, Cooper%
\BCBL {}\ \BBA {} Hauskrecht%
}{%
P.~Naeini%
\ \protect \BOthers {.}}{%
{\protect \APACyear {2015}}%
}]{%
naeini2015}
\APACinsertmetastar {%
naeini2015}%
\begin{APACrefauthors}%
Naeini, P.%
, Cooper, G.F.%
\BCBL {} Hauskrecht, M.%
\end{APACrefauthors}%
\unskip\
\newblock
\APACrefYearMonthDay{2015}{}{}.
\newblock
{\BBOQ}\APACrefatitle {{Obtaining Well Calibrated Probabilities Using Bayesian
  Binning}} {{Obtaining Well Calibrated Probabilities Using Bayesian
  Binning}}.{\BBCQ}
\newblock
 \APACrefbtitle {29th {AAAI} Conference on Artificial Intelligence} {29th
  {AAAI} conference on artificial intelligence}\ (\BVOL~29).
\PrintBackRefs{\CurrentBib}

\bibitem [\protect \citeauthoryear {%
Niculescu-Mizil%
\ \BBA {} Caruana%
}{%
Niculescu-Mizil%
\ \BBA {} Caruana%
}{%
{\protect \APACyear {2005}}%
}]{%
niculescu-mizil2005}
\APACinsertmetastar {%
niculescu-mizil2005}%
\begin{APACrefauthors}%
Niculescu-Mizil, A.%
\BCBT {}\ \BBA {} Caruana, R.%
\end{APACrefauthors}%
\unskip\
\newblock
\APACrefYearMonthDay{2005}{}{}.
\newblock
{\BBOQ}\APACrefatitle {{Predicting good probabilities with supervised
  learning}} {{Predicting good probabilities with supervised learning}}.{\BBCQ}
\newblock
 \APACrefbtitle {22nd International Conference on Machine Learning ({ICML'05})}
  {22nd international conference on machine learning ({ICML'05})}\ (\BPGS\
  625--632).
\newblock
\APACaddressPublisher{New York, New York, USA}{ACM Press}.
\PrintBackRefs{\CurrentBib}

\bibitem [\protect \citeauthoryear {%
Ovadia%
\ \protect \BOthers {.}}{%
Ovadia%
\ \protect \BOthers {.}}{%
{\protect \APACyear {2019}}%
}]{%
ovadia2019can}
\APACinsertmetastar {%
ovadia2019can}%
\begin{APACrefauthors}%
Ovadia, Y.%
, Fertig, E.%
, Ren, J.%
, Nado, Z.%
, Sculley, D.%
, Nowozin, S.%
\BDBL {}Snoek, J.%
\end{APACrefauthors}%
\unskip\
\newblock
\APACrefYearMonthDay{2019}{}{}.
\newblock
{\BBOQ}\APACrefatitle {{Can you trust your model's uncertainty? Evaluating
  predictive uncertainty under dataset shift}} {{Can you trust your model's
  uncertainty? Evaluating predictive uncertainty under dataset shift}}.{\BBCQ}
\newblock
 \APACrefbtitle {Advances in Neural Information Processing Systems 32}
  {Advances in neural information processing systems 32}\ (\BVOL~32).
\PrintBackRefs{\CurrentBib}

\bibitem [\protect \citeauthoryear {%
Patel%
, Beluch%
, Yang%
, Pfeiffer%
\BCBL {}\ \BBA {} Zhang%
}{%
Patel%
\ \protect \BOthers {.}}{%
{\protect \APACyear {2021}}%
}]{%
patel2021multiclass}
\APACinsertmetastar {%
patel2021multiclass}%
\begin{APACrefauthors}%
Patel, K.%
, Beluch, W.H.%
, Yang, B.%
, Pfeiffer, M.%
\BCBL {} Zhang, D.%
\end{APACrefauthors}%
\unskip\
\newblock
\APACrefYearMonthDay{2021}{}{}.
\newblock
{\BBOQ}\APACrefatitle {Multi-Class Uncertainty Calibration via Mutual
  Information Maximization-based Binning} {Multi-class uncertainty calibration
  via mutual information maximization-based binning}.{\BBCQ}
\newblock
 \APACrefbtitle {International Conference on Learning Representations.}
  {International conference on learning representations.}
\newblock
\begin{APACrefURL} {https://openreview.net/forum?id=AICNpd8ke-m}
  \end{APACrefURL}
\PrintBackRefs{\CurrentBib}

\bibitem [\protect \citeauthoryear {%
Perello-Nieto%
, Song%
, {Silva Filho}%
\BCBL {}\ \BBA {} Kängsepp%
}{%
Perello-Nieto%
\ \protect \BOthers {.}}{%
{\protect \APACyear {2021}}%
}]{%
pycalib}
\APACinsertmetastar {%
pycalib}%
\begin{APACrefauthors}%
Perello-Nieto, M.%
, Song, H.%
, {Silva Filho}, T.%
\BCBL {} Kängsepp, M.%
\end{APACrefauthors}%
\unskip\
\newblock
\APACrefYearMonthDay{2021}{{\APACmonth{08}}}{}.
\newblock
\APACrefbtitle {PyCalib a library for classifier calibration.} {Pycalib a
  library for classifier calibration.}
\newblock
\APACaddressPublisher{}{Zenodo}.
\newblock
\begin{APACrefURL} {https://doi.org/10.5281/zenodo.5518877} \end{APACrefURL}
\newblock
\APACrefnote{If you use this software, please cite it as below.}
\newblock
\begin{APACrefDOI} \doi{10.5281/zenodo.5518877} \end{APACrefDOI}
\PrintBackRefs{\CurrentBib}

\bibitem [\protect \citeauthoryear {%
J.~Platt%
}{%
J.~Platt%
}{%
{\protect \APACyear {2000}}%
}]{%
platt2000}
\APACinsertmetastar {%
platt2000}%
\begin{APACrefauthors}%
Platt, J.%
\end{APACrefauthors}%
\unskip\
\newblock
\APACrefYearMonthDay{2000}{}{}.
\newblock
{\BBOQ}\APACrefatitle {{Probabilities for SV Machines}} {{Probabilities for SV
  Machines}}.{\BBCQ}
\newblock
 A.J.~Smola, P.~Bartlett, B.~Sch{\"{o}}lkopf\BCBL {}\ \BBA {} D.~Schuurmans\
  (\BEDS), \APACrefbtitle {Advances in Large-Margin Classifiers} {Advances in
  large-margin classifiers}\ (\BPGS\ 61--74).
\newblock
\APACaddressPublisher{}{MIT Press}.
\PrintBackRefs{\CurrentBib}

\bibitem [\protect \citeauthoryear {%
J.C.~Platt%
}{%
J.C.~Platt%
}{%
{\protect \APACyear {1999}}%
}]{%
platt99probabilisticoutputs}
\APACinsertmetastar {%
platt99probabilisticoutputs}%
\begin{APACrefauthors}%
Platt, J.C.%
\end{APACrefauthors}%
\unskip\
\newblock
\APACrefYearMonthDay{1999}{}{}.
\newblock
{\BBOQ}\APACrefatitle {Probabilistic Outputs for Support Vector Machines and
  Comparisons to Regularized Likelihood Methods} {Probabilistic outputs for
  support vector machines and comparisons to regularized likelihood
  methods}.{\BBCQ}
\newblock
 \APACrefbtitle {Advances in large margin classifiers} {Advances in large
  margin classifiers}\ (\BPGS\ 61--74).
\newblock
\APACaddressPublisher{}{MIT Press}.
\PrintBackRefs{\CurrentBib}

\bibitem [\protect \citeauthoryear {%
Pleiss%
, Raghavan%
, Wu%
, Kleinberg%
\BCBL {}\ \BBA {} Weinberger%
}{%
Pleiss%
\ \protect \BOthers {.}}{%
{\protect \APACyear {2017}}%
}]{%
pleiss2017fairness}
\APACinsertmetastar {%
pleiss2017fairness}%
\begin{APACrefauthors}%
Pleiss, G.%
, Raghavan, M.%
, Wu, F.%
, Kleinberg, J.%
\BCBL {} Weinberger, K.Q.%
\end{APACrefauthors}%
\unskip\
\newblock
\APACrefYearMonthDay{2017}{}{}.
\newblock
{\BBOQ}\APACrefatitle {On fairness and calibration} {On fairness and
  calibration}.{\BBCQ}
\newblock
\APACjournalVolNumPages{arXiv preprint arXiv:1709.02012}{}{}{}.
\newblock

\newblock

\PrintBackRefs{\CurrentBib}

\bibitem [\protect \citeauthoryear {%
Roelofs%
, Cain%
, Shlens%
\BCBL {}\ \BBA {} Mozer%
}{%
Roelofs%
\ \protect \BOthers {.}}{%
{\protect \APACyear {2021}}%
}]{%
roelofs2021mitigating}
\APACinsertmetastar {%
roelofs2021mitigating}%
\begin{APACrefauthors}%
Roelofs, R.%
, Cain, N.%
, Shlens, J.%
\BCBL {} Mozer, M.C.%
\end{APACrefauthors}%
\unskip\
\newblock
\APACrefYearMonthDay{2021}{}{}.
\newblock
{\BBOQ}\APACrefatitle {Mitigating Bias in Calibration Error Estimation}
  {Mitigating bias in calibration error estimation}.{\BBCQ}
\newblock
\APACjournalVolNumPages{arXiv preprint arXiv:2012.08668}{}{}{}.
\newblock

\newblock

\PrintBackRefs{\CurrentBib}

\bibitem [\protect \citeauthoryear {%
Romano%
, Patterson%
\BCBL {}\ \BBA {} Candes%
}{%
Romano%
\ \protect \BOthers {.}}{%
{\protect \APACyear {2019}}%
}]{%
romano2019conformalized}
\APACinsertmetastar {%
romano2019conformalized}%
\begin{APACrefauthors}%
Romano, Y.%
, Patterson, E.%
\BCBL {} Candes, E.%
\end{APACrefauthors}%
\unskip\
\newblock
\APACrefYearMonthDay{2019}{}{}.
\newblock
{\BBOQ}\APACrefatitle {Conformalized quantile regression} {Conformalized
  quantile regression}.{\BBCQ}
\newblock
\APACjournalVolNumPages{Advances in Neural Information Processing
  Systems}{32}{}{3543--3553}.
\newblock

\newblock

\PrintBackRefs{\CurrentBib}

\bibitem [\protect \citeauthoryear {%
Sahoo%
, Zhao%
, Chen%
\BCBL {}\ \BBA {} Ermon%
}{%
Sahoo%
\ \protect \BOthers {.}}{%
{\protect \APACyear {2021}}%
}]{%
sahoo3reliable}
\APACinsertmetastar {%
sahoo3reliable}%
\begin{APACrefauthors}%
Sahoo, R.%
, Zhao, S.%
, Chen, A.%
\BCBL {} Ermon, S.%
\end{APACrefauthors}%
\unskip\
\newblock
\APACrefYearMonthDay{2021}{}{}.
\newblock
{\BBOQ}\APACrefatitle {Reliable Decisions with Threshold Calibration} {Reliable
  decisions with threshold calibration}.{\BBCQ}
\newblock
\APACjournalVolNumPages{Threshold}{3}{}{3--5}.
\newblock

\newblock

\PrintBackRefs{\CurrentBib}

\bibitem [\protect \citeauthoryear {%
Sanders%
}{%
Sanders%
}{%
{\protect \APACyear {1963}}%
}]{%
sanders1963}
\APACinsertmetastar {%
sanders1963}%
\begin{APACrefauthors}%
Sanders, F.%
\end{APACrefauthors}%
\unskip\
\newblock
\APACrefYearMonthDay{1963}{apr}{}.
\newblock
{\BBOQ}\APACrefatitle {{On Subjective Probability Forecasting}} {{On Subjective
  Probability Forecasting}}.{\BBCQ}
\newblock
\APACjournalVolNumPages{Journal of Applied Meteorology}{2}{2}{191--201}.
\newblock

\newblock

\PrintBackRefs{\CurrentBib}

\bibitem [\protect \citeauthoryear {%
Savage%
}{%
Savage%
}{%
{\protect \APACyear {1971}}%
}]{%
savage1971}
\APACinsertmetastar {%
savage1971}%
\begin{APACrefauthors}%
Savage, L.J.%
\end{APACrefauthors}%
\unskip\
\newblock
\APACrefYearMonthDay{1971}{}{}.
\newblock
{\BBOQ}\APACrefatitle {Elicitation of personal probabilities and expectations}
  {Elicitation of personal probabilities and expectations}.{\BBCQ}
\newblock
\APACjournalVolNumPages{Journal of the American Statistical
  Association}{66}{336}{783--801}.
\newblock

\newblock

\PrintBackRefs{\CurrentBib}

\bibitem [\protect \citeauthoryear {%
Senge%
\ \protect \BOthers {.}}{%
Senge%
\ \protect \BOthers {.}}{%
{\protect \APACyear {2014}}%
}]{%
senge2014reliable}
\APACinsertmetastar {%
senge2014reliable}%
\begin{APACrefauthors}%
Senge, R.%
, B{\"o}sner, S.%
, Dembczy{\'n}ski, K.%
, Haasenritter, J.%
, Hirsch, O.%
, Donner-Banzhoff, N.%
\BCBL {} H{\"u}llermeier, E.%
\end{APACrefauthors}%
\unskip\
\newblock
\APACrefYearMonthDay{2014}{}{}.
\newblock
{\BBOQ}\APACrefatitle {Reliable classification: Learning classifiers that
  distinguish aleatoric and epistemic uncertainty} {Reliable classification:
  Learning classifiers that distinguish aleatoric and epistemic
  uncertainty}.{\BBCQ}
\newblock
\APACjournalVolNumPages{Information Sciences}{255}{}{16--29}.
\newblock

\newblock

\PrintBackRefs{\CurrentBib}

\bibitem [\protect \citeauthoryear {%
Shaker%
\ \BBA {} Hüllermeier%
}{%
Shaker%
\ \BBA {} Hüllermeier%
}{%
{\protect \APACyear {2021}}%
}]{%
shaker2021ensemble}
\APACinsertmetastar {%
shaker2021ensemble}%
\begin{APACrefauthors}%
Shaker, M.H.%
\BCBT {}\ \BBA {} Hüllermeier, E.%
\end{APACrefauthors}%
\unskip\
\newblock
\APACrefYearMonthDay{2021}{}{}.
\newblock
\APACrefbtitle {Ensemble-based Uncertainty Quantification: Bayesian versus
  Credal Inference.} {Ensemble-based uncertainty quantification: Bayesian
  versus credal inference.}
\newblock
\APACaddressPublisher{}{arXiv}.
\newblock
\begin{APACrefURL} {https://arxiv.org/abs/2107.10384} \end{APACrefURL}
\newblock
\begin{APACrefDOI} \doi{10.48550/ARXIV.2107.10384} \end{APACrefDOI}
\PrintBackRefs{\CurrentBib}

\bibitem [\protect \citeauthoryear {%
Song%
, Diethe%
, Kull%
\BCBL {}\ \BBA {} Flach%
}{%
Song%
\ \protect \BOthers {.}}{%
{\protect \APACyear {2019}}%
}]{%
pmlr-v97-song19a}
\APACinsertmetastar {%
pmlr-v97-song19a}%
\begin{APACrefauthors}%
Song, H.%
, Diethe, T.%
, Kull, M.%
\BCBL {} Flach, P.%
\end{APACrefauthors}%
\unskip\
\newblock
\APACrefYearMonthDay{2019}{09--15 Jun}{}.
\newblock
{\BBOQ}\APACrefatitle {Distribution calibration for regression} {Distribution
  calibration for regression}.{\BBCQ}
\newblock
 K.~Chaudhuri\ \BBA {} R.~Salakhutdinov\ (\BEDS), \APACrefbtitle {36th
  International Conference on Machine Learning} {36th international conference
  on machine learning}\ (\BVOL~97, \BPGS\ 5897--5906).
\newblock
\APACaddressPublisher{Long Beach, California, USA}{PMLR}.
\PrintBackRefs{\CurrentBib}

\bibitem [\protect \citeauthoryear {%
Thulasidasan%
, Chennupati%
, Bilmes%
, Bhattacharya%
\BCBL {}\ \BBA {} Michalak%
}{%
Thulasidasan%
\ \protect \BOthers {.}}{%
{\protect \APACyear {2019}}%
}]{%
thulasidasan2019mixup}
\APACinsertmetastar {%
thulasidasan2019mixup}%
\begin{APACrefauthors}%
Thulasidasan, S.%
, Chennupati, G.%
, Bilmes, J.%
, Bhattacharya, T.%
\BCBL {} Michalak, S.%
\end{APACrefauthors}%
\unskip\
\newblock
\APACrefYearMonthDay{2019}{}{}.
\newblock
{\BBOQ}\APACrefatitle {On mixup training: Improved calibration and predictive
  uncertainty for deep neural networks} {On mixup training: Improved
  calibration and predictive uncertainty for deep neural networks}.{\BBCQ}
\newblock
\APACjournalVolNumPages{arXiv preprint arXiv:1905.11001}{}{}{}.
\newblock

\newblock

\PrintBackRefs{\CurrentBib}

\bibitem [\protect \citeauthoryear {%
Tran%
, Bonilla%
, Cunningham%
, Michiardi%
\BCBL {}\ \BBA {} Filippone%
}{%
Tran%
\ \protect \BOthers {.}}{%
{\protect \APACyear {2019}}%
}]{%
tran2019calibrating}
\APACinsertmetastar {%
tran2019calibrating}%
\begin{APACrefauthors}%
Tran, G\BHBI L.%
, Bonilla, E.V.%
, Cunningham, J.%
, Michiardi, P.%
\BCBL {} Filippone, M.%
\end{APACrefauthors}%
\unskip\
\newblock
\APACrefYearMonthDay{2019}{}{}.
\newblock
{\BBOQ}\APACrefatitle {Calibrating deep convolutional gaussian processes}
  {Calibrating deep convolutional gaussian processes}.{\BBCQ}
\newblock
 \APACrefbtitle {The 22nd International Conference on Artificial Intelligence
  and Statistics} {The 22nd international conference on artificial intelligence
  and statistics}\ (\BPGS\ 1554--1563).
\PrintBackRefs{\CurrentBib}

\bibitem [\protect \citeauthoryear {%
Vaicenavicius%
\ \protect \BOthers {.}}{%
Vaicenavicius%
\ \protect \BOthers {.}}{%
{\protect \APACyear {2019}}%
}]{%
vaicenavicius2019evaluating}
\APACinsertmetastar {%
vaicenavicius2019evaluating}%
\begin{APACrefauthors}%
Vaicenavicius, J.%
, Widmann, D.%
, Andersson, C.%
, Lindsten, F.%
, Roll, J.%
\BCBL {} Sch{\"o}n, T.%
\end{APACrefauthors}%
\unskip\
\newblock
\APACrefYearMonthDay{2019}{}{}.
\newblock
{\BBOQ}\APACrefatitle {Evaluating model calibration in classification}
  {Evaluating model calibration in classification}.{\BBCQ}
\newblock
 \APACrefbtitle {The 22nd International Conference on Artificial Intelligence
  and Statistics} {The 22nd international conference on artificial intelligence
  and statistics}\ (\BPGS\ 3459--3467).
\PrintBackRefs{\CurrentBib}

\bibitem [\protect \citeauthoryear {%
Vovk%
, Gammerman%
\BCBL {}\ \BBA {} Shafer%
}{%
Vovk%
\ \protect \BOthers {.}}{%
{\protect \APACyear {2005}}%
}]{%
vovk2005algorithmic}
\APACinsertmetastar {%
vovk2005algorithmic}%
\begin{APACrefauthors}%
Vovk, V.%
, Gammerman, A.%
\BCBL {} Shafer, G.%
\end{APACrefauthors}%
\unskip\
\newblock
\APACrefYear{2005}.
\newblock
\APACrefbtitle {Algorithmic learning in a random world} {Algorithmic learning
  in a random world}.
\newblock
\APACaddressPublisher{}{Springer Science \& Business Media}.
\PrintBackRefs{\CurrentBib}

\bibitem [\protect \citeauthoryear {%
Vovk%
\ \BBA {} Petej%
}{%
Vovk%
\ \BBA {} Petej%
}{%
{\protect \APACyear {2012}}%
}]{%
vovk2012venn}
\APACinsertmetastar {%
vovk2012venn}%
\begin{APACrefauthors}%
Vovk, V.%
\BCBT {}\ \BBA {} Petej, I.%
\end{APACrefauthors}%
\unskip\
\newblock
\APACrefYearMonthDay{2012}{}{}.
\newblock
{\BBOQ}\APACrefatitle {Venn-abers predictors} {Venn-abers predictors}.{\BBCQ}
\newblock
\APACjournalVolNumPages{arXiv preprint arXiv:1211.0025}{}{}{}.
\newblock

\newblock

\PrintBackRefs{\CurrentBib}

\bibitem [\protect \citeauthoryear {%
Vovk%
, Shafer%
\BCBL {}\ \BBA {} Nouretdinov%
}{%
Vovk%
\ \protect \BOthers {.}}{%
{\protect \APACyear {2003}}%
}]{%
vovk2003self}
\APACinsertmetastar {%
vovk2003self}%
\begin{APACrefauthors}%
Vovk, V.%
, Shafer, G.%
\BCBL {} Nouretdinov, I.%
\end{APACrefauthors}%
\unskip\
\newblock
\APACrefYearMonthDay{2003}{}{}.
\newblock
{\BBOQ}\APACrefatitle {Self-calibrating probability forecasting}
  {Self-calibrating probability forecasting}.{\BBCQ}
\newblock
\APACjournalVolNumPages{Advances in neural information processing
  systems}{16}{}{}.
\newblock

\newblock

\PrintBackRefs{\CurrentBib}

\bibitem [\protect \citeauthoryear {%
Wang%
, Li%
\BCBL {}\ \BBA {} Dang%
}{%
Wang%
\ \protect \BOthers {.}}{%
{\protect \APACyear {2019}}%
}]{%
wang2019calibrating}
\APACinsertmetastar {%
wang2019calibrating}%
\begin{APACrefauthors}%
Wang, Y.%
, Li, L.%
\BCBL {} Dang, C.%
\end{APACrefauthors}%
\unskip\
\newblock
\APACrefYearMonthDay{2019}{}{}.
\newblock
{\BBOQ}\APACrefatitle {Calibrating classification probabilities with
  shape-restricted polynomial regression} {Calibrating classification
  probabilities with shape-restricted polynomial regression}.{\BBCQ}
\newblock
\APACjournalVolNumPages{IEEE transactions on pattern analysis and machine
  intelligence}{41}{8}{1813--1827}.
\newblock

\newblock

\PrintBackRefs{\CurrentBib}

\bibitem [\protect \citeauthoryear {%
Wen%
, Tran%
\BCBL {}\ \BBA {} Ba%
}{%
Wen%
\ \protect \BOthers {.}}{%
{\protect \APACyear {2020}}%
}]{%
wen2020batchensemble}
\APACinsertmetastar {%
wen2020batchensemble}%
\begin{APACrefauthors}%
Wen, Y.%
, Tran, D.%
\BCBL {} Ba, J.%
\end{APACrefauthors}%
\unskip\
\newblock
\APACrefYearMonthDay{2020}{}{}.
\newblock
{\BBOQ}\APACrefatitle {Batchensemble: an alternative approach to efficient
  ensemble and lifelong learning} {Batchensemble: an alternative approach to
  efficient ensemble and lifelong learning}.{\BBCQ}
\newblock
\APACjournalVolNumPages{arXiv preprint arXiv:2002.06715}{}{}{}.
\newblock

\newblock

\PrintBackRefs{\CurrentBib}

\bibitem [\protect \citeauthoryear {%
Wenger%
, Kjellstr{\"o}m%
\BCBL {}\ \BBA {} Triebel%
}{%
Wenger%
\ \protect \BOthers {.}}{%
{\protect \APACyear {2020}}%
}]{%
wenger2020}
\APACinsertmetastar {%
wenger2020}%
\begin{APACrefauthors}%
Wenger, J.%
, Kjellstr{\"o}m, H.%
\BCBL {} Triebel, R.%
\end{APACrefauthors}%
\unskip\
\newblock
\APACrefYearMonthDay{2020}{}{}.
\newblock
{\BBOQ}\APACrefatitle {Non-parametric calibration for classification}
  {Non-parametric calibration for classification}.{\BBCQ}
\newblock
 \APACrefbtitle {International Conference on Artificial Intelligence and
  Statistics} {International conference on artificial intelligence and
  statistics}\ (\BPGS\ 178--190).
\PrintBackRefs{\CurrentBib}

\bibitem [\protect \citeauthoryear {%
Widmann%
, Lindsten%
\BCBL {}\ \BBA {} Zachariah%
}{%
Widmann%
\ \protect \BOthers {.}}{%
{\protect \APACyear {2019}}%
}]{%
widmann2019calibration}
\APACinsertmetastar {%
widmann2019calibration}%
\begin{APACrefauthors}%
Widmann, D.%
, Lindsten, F.%
\BCBL {} Zachariah, D.%
\end{APACrefauthors}%
\unskip\
\newblock
\APACrefYearMonthDay{2019}{}{}.
\newblock
{\BBOQ}\APACrefatitle {Calibration tests in multi-class classification: A
  unifying framework} {Calibration tests in multi-class classification: A
  unifying framework}.{\BBCQ}
\newblock
 \APACrefbtitle {Advances in Neural Information Processing Systems} {Advances
  in neural information processing systems}\ (\BPGS\ 12257--12267).
\PrintBackRefs{\CurrentBib}

\bibitem [\protect \citeauthoryear {%
Widmann%
, Lindsten%
\BCBL {}\ \BBA {} Zachariah%
}{%
Widmann%
\ \protect \BOthers {.}}{%
{\protect \APACyear {2022}}%
}]{%
widmann2022calibration}
\APACinsertmetastar {%
widmann2022calibration}%
\begin{APACrefauthors}%
Widmann, D.%
, Lindsten, F.%
\BCBL {} Zachariah, D.%
\end{APACrefauthors}%
\unskip\
\newblock
\APACrefYearMonthDay{2022}{}{}.
\newblock
\APACrefbtitle {Calibration tests beyond classification.} {Calibration tests
  beyond classification.}
\newblock
\APACaddressPublisher{}{arXiv}.
\newblock
\begin{APACrefURL} {https://arxiv.org/abs/2210.13355} \end{APACrefURL}
\newblock
\begin{APACrefDOI} \doi{10.48550/ARXIV.2210.13355} \end{APACrefDOI}
\PrintBackRefs{\CurrentBib}

\bibitem [\protect \citeauthoryear {%
Winkler%
}{%
Winkler%
}{%
{\protect \APACyear {1969}}%
}]{%
winkler1969scoring}
\APACinsertmetastar {%
winkler1969scoring}%
\begin{APACrefauthors}%
Winkler, R.L.%
\end{APACrefauthors}%
\unskip\
\newblock
\APACrefYearMonthDay{1969}{}{}.
\newblock
{\BBOQ}\APACrefatitle {Scoring Rules and the Evaluation of Probability
  Assessors} {Scoring rules and the evaluation of probability
  assessors}.{\BBCQ}
\newblock
\APACjournalVolNumPages{Journal of the American Statistical
  Association}{64}{327}{1073--1078}.
\newblock

\newblock

\PrintBackRefs{\CurrentBib}

\bibitem [\protect \citeauthoryear {%
Zadrozny%
\ \BBA {} Elkan%
}{%
Zadrozny%
\ \BBA {} Elkan%
}{%
{\protect \APACyear {2001}}%
}]{%
zadrozny2001obtaining}
\APACinsertmetastar {%
zadrozny2001obtaining}%
\begin{APACrefauthors}%
Zadrozny, B.%
\BCBT {}\ \BBA {} Elkan, C.%
\end{APACrefauthors}%
\unskip\
\newblock
\APACrefYearMonthDay{2001}{}{}.
\newblock
{\BBOQ}\APACrefatitle {Obtaining calibrated probability estimates from decision
  trees and naive bayesian classifiers} {Obtaining calibrated probability
  estimates from decision trees and naive bayesian classifiers}.{\BBCQ}
\newblock
 \APACrefbtitle {{ICML}} {{ICML}}\ (\BVOL~1, \BPGS\ 609--616).
\PrintBackRefs{\CurrentBib}

\bibitem [\protect \citeauthoryear {%
Zadrozny%
\ \BBA {} Elkan%
}{%
Zadrozny%
\ \BBA {} Elkan%
}{%
{\protect \APACyear {2002}}%
}]{%
zadrozny2002}
\APACinsertmetastar {%
zadrozny2002}%
\begin{APACrefauthors}%
Zadrozny, B.%
\BCBT {}\ \BBA {} Elkan, C.%
\end{APACrefauthors}%
\unskip\
\newblock
\APACrefYearMonthDay{2002}{}{}.
\newblock
{\BBOQ}\APACrefatitle {{Transforming Classifier Scores into Accurate Multiclass
  Probability Estimates}} {{Transforming Classifier Scores into Accurate
  Multiclass Probability Estimates}}.{\BBCQ}
\newblock
 \APACrefbtitle {8th {ACM SIGKDD} international conference on Knowledge
  discovery and data mining - {KDD} '02} {8th {ACM SIGKDD} international
  conference on knowledge discovery and data mining - {KDD} '02}\ (\BPGS\
  694--699).
\newblock
\APACaddressPublisher{New York, New York, USA}{ACM Press}.
\PrintBackRefs{\CurrentBib}

\bibitem [\protect \citeauthoryear {%
Zhang%
, Kailkhura%
\BCBL {}\ \BBA {} Han%
}{%
Zhang%
\ \protect \BOthers {.}}{%
{\protect \APACyear {2020}}%
}]{%
zhang2020mix}
\APACinsertmetastar {%
zhang2020mix}%
\begin{APACrefauthors}%
Zhang, J.%
, Kailkhura, B.%
\BCBL {} Han, T.Y\BHBI J.%
\end{APACrefauthors}%
\unskip\
\newblock
\APACrefYearMonthDay{2020}{}{}.
\newblock
{\BBOQ}\APACrefatitle {Mix-n-match: Ensemble and compositional methods for
  uncertainty calibration in deep learning} {Mix-n-match: Ensemble and
  compositional methods for uncertainty calibration in deep learning}.{\BBCQ}
\newblock
 \APACrefbtitle {International Conference on Machine Learning} {International
  conference on machine learning}\ (\BPGS\ 11117--11128).
\PrintBackRefs{\CurrentBib}

\end{thebibliography}

\end{document}